\documentclass{article}

\usepackage{arxiv}

\usepackage[utf8]{inputenc} 
\usepackage[T1]{fontenc}    
\usepackage{hyperref}       
\usepackage{url}            
\usepackage{booktabs}       
\usepackage{amsfonts}       
\usepackage{nicefrac}       
\usepackage{microtype}      
\usepackage{lipsum}		
\usepackage{graphicx}
\usepackage{natbib}
\usepackage{doi}
\usepackage{svg}
\usepackage{xcolor}
\usepackage{comment}
\usepackage{float}
\usepackage{amsmath} 
\usepackage{placeins}
\usepackage{adjustbox}
\usepackage{amsmath}

\title{Simulation of  a first prototypical 3D solution for Indoor 
Localization based on Directed and Reflected Signals}

\author{\href{https://orcid.org/0000-0001-7881-4563}{\includegraphics[scale=0.06]{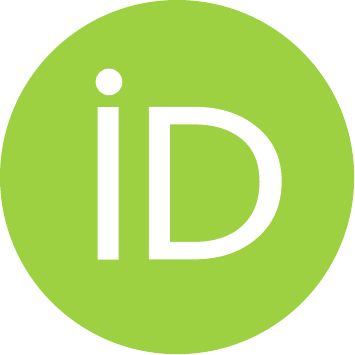}\hspace{1mm}Sneha Mohanty}\\
	Computer Networks and Telematics\\
	University of Freiburg\\
	Freiburg, Germany\\
	\texttt{mohanty@informatik.uni-freiburg.de} \\
	\And
	{Milan Müller}\\
	Computer Networks and Telematics\\
	University of Freiburg\\
	Freiburg, Germany\\
	\texttt{milan.mueller@students.uni-freiburg.de} 
    \And
	\href{https://orcid.org/0000-0002-8320-8581}{\includegraphics[scale=0.06]{orcid.pdf}\hspace{1mm}Christian Schindelhauer} \\Computer Networks and Telematics\\
	University of Freiburg\\
	Freiburg, Germany\\
	\texttt{schindel@informatik.uni-freiburg.de}
 \\
}

\renewcommand{\headeright}{Technical Report}
\renewcommand{\undertitle}{Technical Report}
\renewcommand{\shorttitle}{3D solution for Indoor Localization based on Directed and Reflected Signals}

\hypersetup{
pdftitle={A template for the arxiv style},
pdfsubject={q-bio.NC, q-bio.QM},
pdfauthor={David S.~Hippocampus, Elias D.~Striatum},
pdfkeywords={First keyword, Second keyword, More},
}

\begin{document}
\maketitle

\begin{abstract}\label{abs}
We introduce a solution for a specific case of Indoor Localization which involves a directed signal, a reflected signal from the wall and the time difference between them. This solution includes robust localization with a given wall, finding the right wall from a group of walls, obtaining the reflecting wall from measurements, using averaging techniques for improving measurements with errors and successfully grouping measurements regarding reflecting walls. It also includes performing self-calibration by computation of wall distance and direction introducing algorithms such as All pairs, Disjoint pairs and Overlapping pairs and clustering walls based on Inversion and Gnomonic Projection. Several of these algorithms are then compared in order to ameliorate the effects of measurement errors.
\end{abstract}


\section{Introduction}\label{intro}
Our solution localizes indoor objects using zero initialization, self-calibration and mapping of the room.

 \subsection{Motivation}\label{motivation}
Previous works in this area of research have used triangulation and trilateration techniques in which the source was at the same location as the receiver. This did not compute the correct walls in the end. To address this issue, the Indoor Localization of Directed and Reflected Signals (ILDARS) approach proposes a mathematical model separating the source from the receiver.
Using the information of the direction of the actual signal, reflection from the wall and the time difference between these, one can then compute the wall distance using various techniques as mentioned in subsequent sections of the paper.

   \subsection{Problem Description}\label{pb1}
   
We refer to our problem as Indoor Localization of Directed and Reflected Signals (ILDARS). Given $n$-Line of Sight order one reflection and time difference of arrival measurements (abbreviated from hereon as LOS-T1R-TDoA), we need to compute all sources such as $S_{1}, S_{2},...., S_{n}$. Assuming that the number of walls is $n$ and that atleast two sources reflect from the same wall, it is possible to solve the problem for two exact inputs in general position in 3D as well as the problem for two exact inputs in general position in 2D. This makes a straight-forward generalization with quadratic run-time possible. The practical problem however remains to find a stable solution for real-world inputs for random sources in 3D. 

\section{Related Work}\label{relwork}

\cite{6112205} proposes a measure to compute the distance to a wall. \cite{dokmanic2013acoustic} writes that a single snap and many microphones as well as reflections from these would be sufficient to reveal the shape of a room. \cite{plumbey13} involves a similar observation as here but without directions. \cite{Parhizkar2014SinglechannelIM} comes up with a reverberation based model using Single-channel indoor microphone localization. \cite{al2014SourceLA} proposes source localization and tracking in non-convex rooms.

\section{ILDARS for 3D}\label{ildars}

The Indoor Localization Directed and Reflected Signals (ILDARS) approach includes input signals coming in, localization based on reception which involves real-time part with same inputs and localization using different walls and a self-calibration part. The self-calibration aspect includes clustering and computation of walls. The output of this consists of position of sound source.

\subsection{Problem Description}\label{pb2}

Assume two general positioned sources in 3D, one reflecting wall with LOS-T1R-TDoA measurements. Given the directions, $\vec{v_{1}}$ and $\vec{v_{2}}$ of two signals, the directions $\vec{w_{1}}$ and $\vec{w_{2}}$ of their reflections from the same wall, the time delays of signals and reflections, $\Delta_{1}$ and $\Delta_{2}$. Compute the distances, $p_{1}$ and $p_{2}$ from signals to the origin.

\subsection{Proposed solution}\label{psol}
If $\vec{v_{1}}$, $\vec{v_{2}}$, $\vec{w_{1}}$ and $\vec{w_{2}}$ are not co-planar, compute intersecting direction: 
\begin{align}\label{vecudef}
    \vec{u} = \dfrac{(\vec{v_{1}} \times \vec{w_{1}}) \times (\vec{v_{2}} \times \vec{w_{2}})}{\mid (\vec{v_{1}} \times \vec{w_{1}}) \times (\vec{v_{2}} \times \vec{w_{2}})  \mid}
\end{align} 
and normal vector to the wall, $\vec{n} = d\vec{u}$. The term $d$ is the distance between the origin and the wall, following the same construction as in \mbox{Fig.~\ref{case_wall_dir}}.

The $\vec{b}$ for the case where $\vec{u} \neq \pm \vec{w}$ is a vector of unspecified length and inside the plane given by $\vec{u}$ and $\vec{w}$, parallel to the reflecting plane. Since we simulate real-world data, we assume all points and directions in general positions.

For $i = 1,2$,
\begin{align} \label{eqb}
\vec{b_{i}} = (\vec{u_{i}} \times \vec{w_{i}}) \times \vec{u}
\end{align}
\begin{align} \label{eqpvector}
p_{i} = \dfrac{\Delta_{i}\vec{w_{i}}\vec{b_{i}}}{((\vec{w_{i}} - \vec{v_{i}}) \cdot \vec{b})}\\
\end{align}
\begin{align} \label{eqsvecgeneric}
\vec{s_{i}} = p_{i}\vec{v_{i}}
\end{align}

\begin{figure}[htbp]
\centering
\includegraphics[width=1.0\columnwidth]{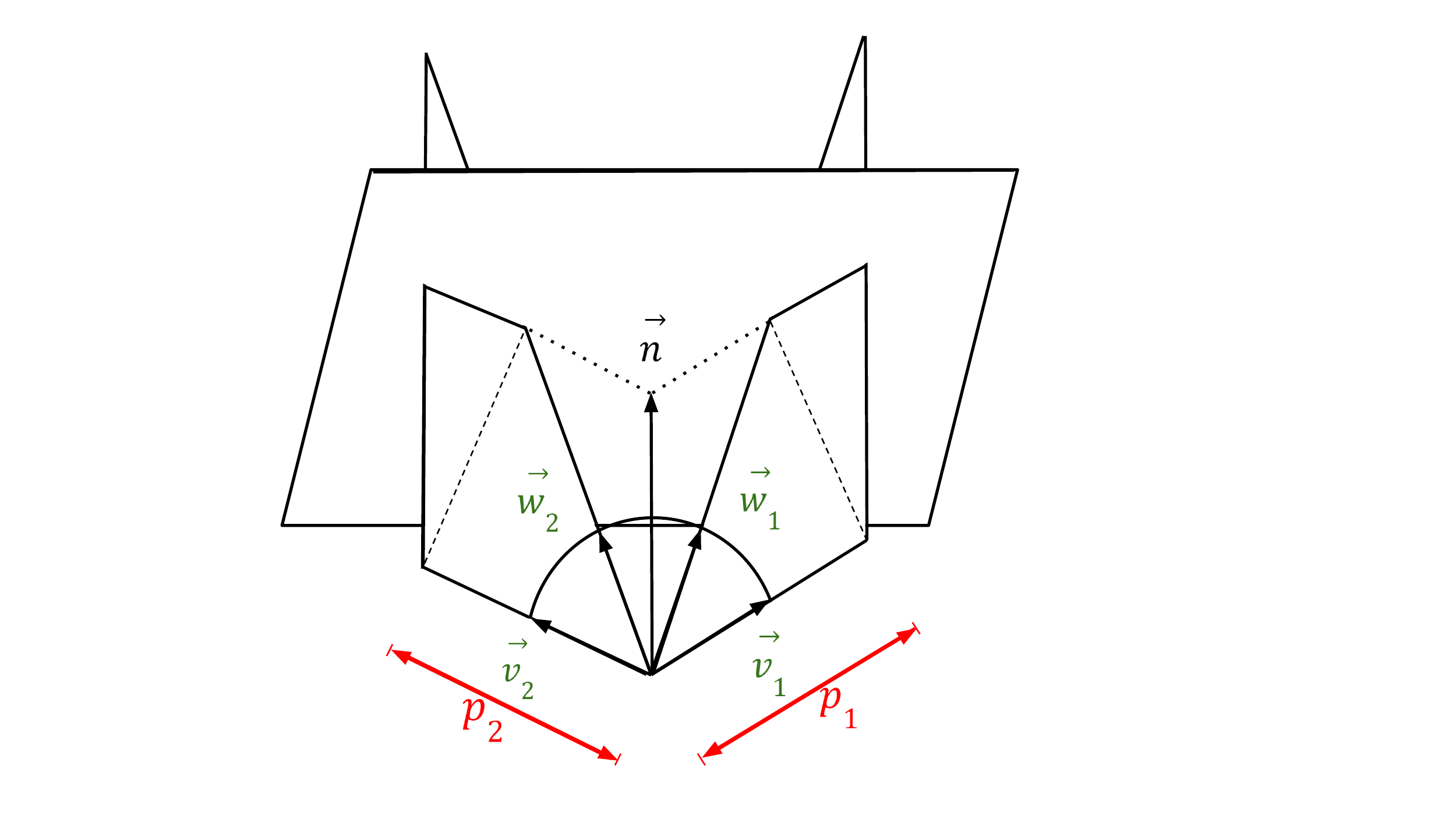}
\caption{Indoor Localization Directed and Reflected Signals
(ILDARS) 3D approach}
\label{case_wherewaldo-3d}
\end{figure}

\FloatBarrier

\subsection{Conclusions for this approach}\label{conc1}

Testing out all combinations of LOS-T1R-TDoA pair leads to conflicts including construction of false positive walls, multiple walls for a reflection and replicated walls because of erroneous inputs. The conclusion of ILDARS for 3D is that it does not scale and hence does not compute the required distances, $p_{1}$ and $p_{2}$.

\section{Design for ILDARS Prototype}\label{desildars}

\begin{figure}[htbp]
\centering
\includegraphics[width=0.5\columnwidth]{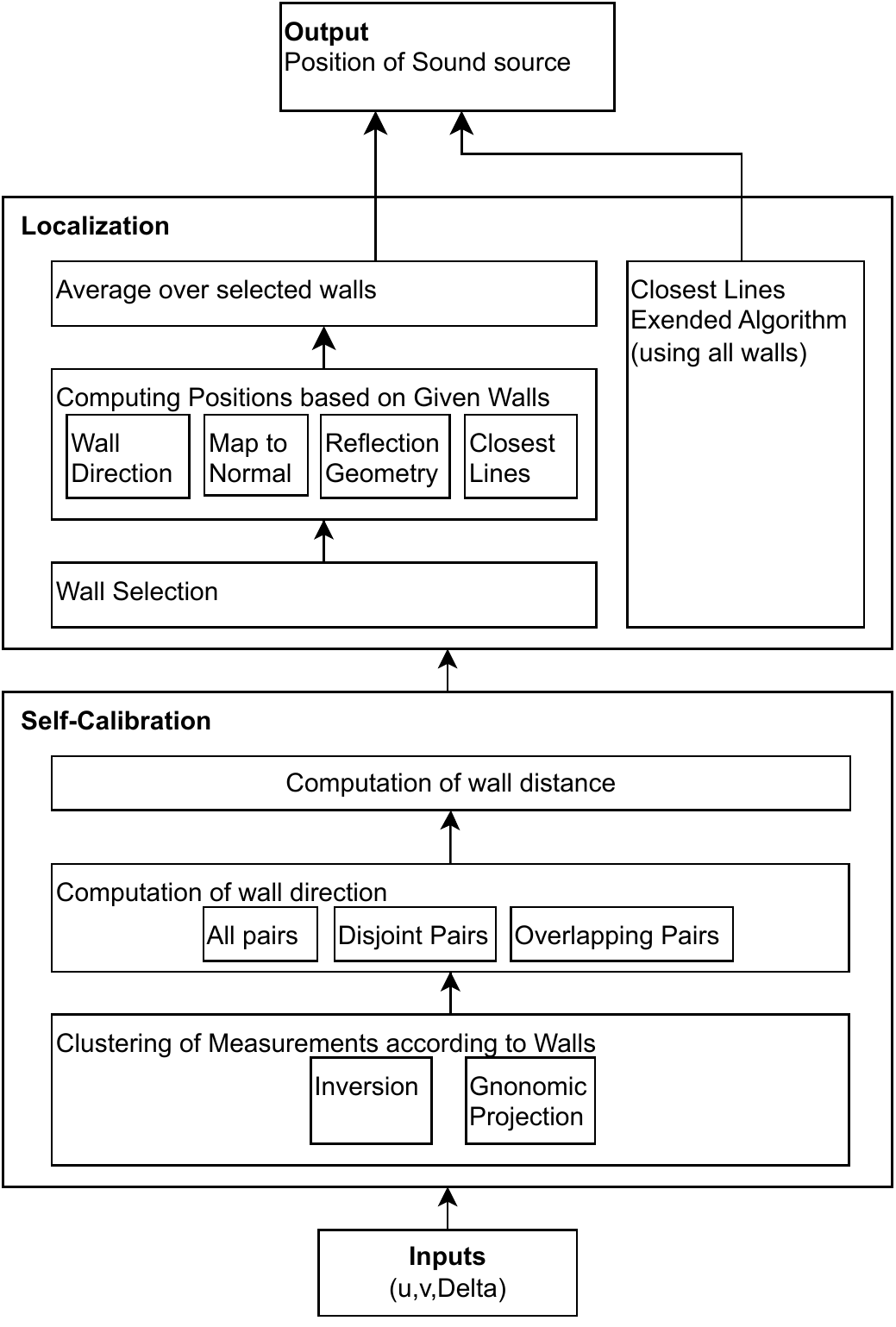}
\caption{Design of the ILDARS system}
\label{case_block}
\end{figure}

\begin{figure}[htbp] 
\centering
\includegraphics[width=0.9\columnwidth]{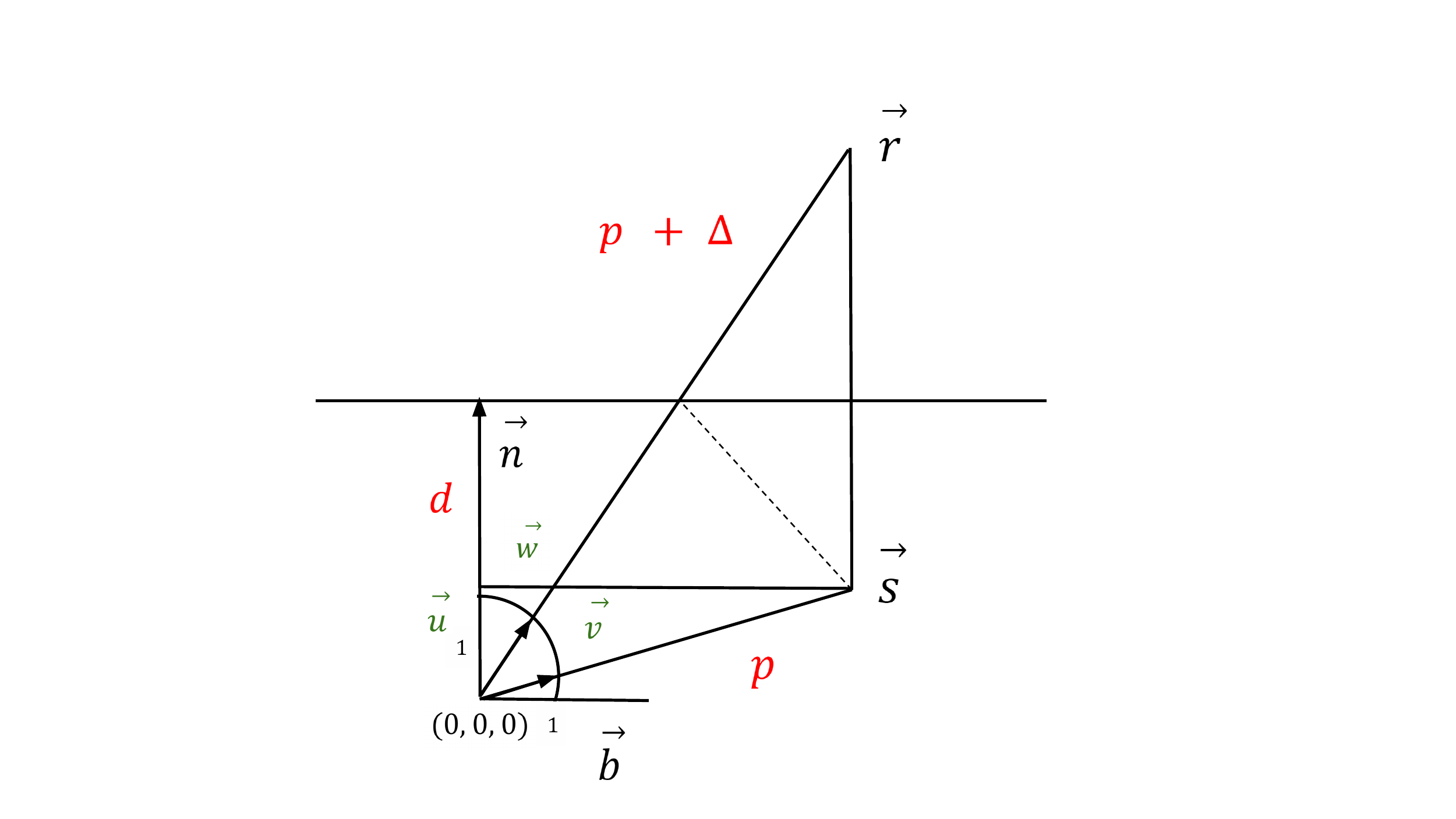} 
\caption{Visualization of the \textit{Wall Direction} algorithm}
\label{case_wall_dir}
\end{figure}

For this paper, we define a \textit{wall} as a simple polygon in a plane and
a \textit{room} as a collection of walls.
We then denote the position of the listener device as $\vec{o}\in{R}^{3}$.
Further, we need at least two sound sources $\vec{s_1},\vec{s_2},\dots,\vec{s_n}\in{R}^{3}$.
The sound sources then emit sound signals which the listener receives and then computes measurements from.
\begin{figure}[htbp]
    \centering
    \includegraphics[width=0.9\textwidth]{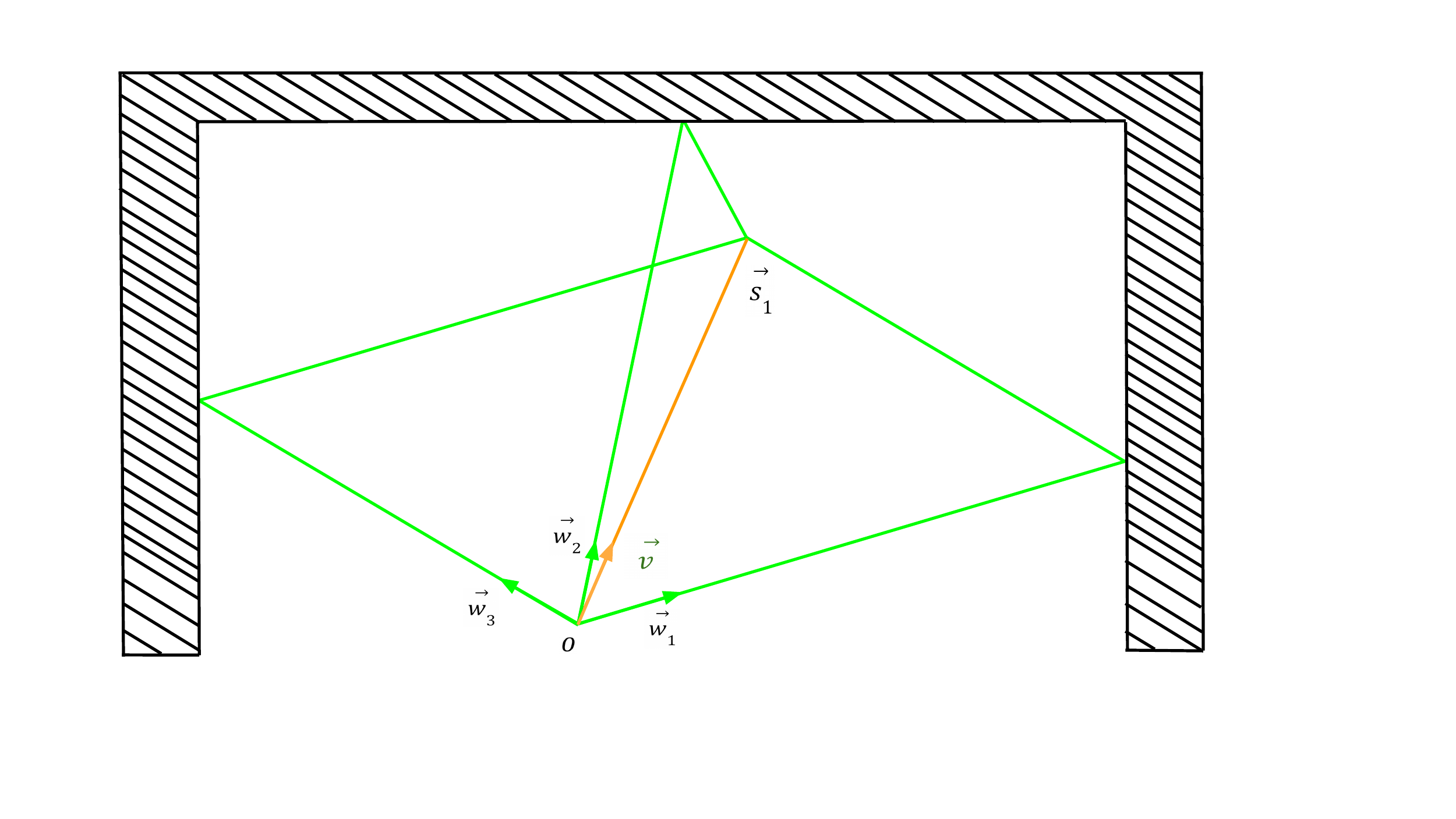}
    \caption{Receiver device at position $o$ receives one direct and multiple reflected signals from sound source $s_1$}
    \label{fig:my_label}
\end{figure}
\FloatBarrier
A measurement is a triple $(\vec{v}, \vec{w}, \Delta)$ where $\vec{v}$ is the direction of a direct signal, $\vec{w}$ is the direction of a reflected signal, which originates from the same sound source as $\vec{v}$, and $\Delta$ is the time difference between receiving the direct and the reflected signal as shown in \mbox{Fig.~\ref{case_wall_dir}}.Our goal is to compute the position of the sound sources $\vec{s_1},\vec{s_2},\dots,\vec{s_n}$.
Note that, because all measurement are relative to the sender, we assume $\vec{o}=0$ in the setup shown here in \mbox{Fig.~\ref{fig:my_label}}.\\
\mbox{Fig.~\ref{case_block}} shows an overview of the design of the ILDARS system: The first step is to find clusters of measurements, each corresponding to one wall, such that all reflected signals in one cluster are reflected from the same wall. For this clustering step we introduce two alternatives titled \textit{Inversion} and \textit{Gnomonic Projection}.\\
After the measurements have been clustered, we can use one cluster of measurements to compute the direction of the corresponding wall. Once the wall's direction is known, we can also compute the distance. For computing the direction of the wall, we need to average over selected pairs of measurements. We have three options for selecting such pairs, namely \textit{All Pairs}, \textit{Disjoint Pairs} and \textit{Overlapping Pairs}.\\
Clustering the measurements and computing the direction and distance of the walls is what we refer to as the Self-Calibration step.
Using the computed wall positions, we can then compute the position of the sender for a given measurement in the Localization step.
The reason for separating the process into these two steps is that the Self-Calibration step could be executed once, and then it's results could be used for new individual measurements in an online variant of the localization step.

\FloatBarrier

\section{Self Calibration}\label{selfcal}
As our initial input, we get a large set of measurements from different walls. 
In the self calibration phase, we need to separate these measurements by the walls that the respective reflections were reflected from.
In addition to that, we also compute the direction and distance of the wall.

\subsection{Clustering}\label{clustering}

\subsubsection{Inversion Approach}\label{inversion}
\begin{figure}[htbp]
    \centering
    \includegraphics[width=0.3\columnwidth]{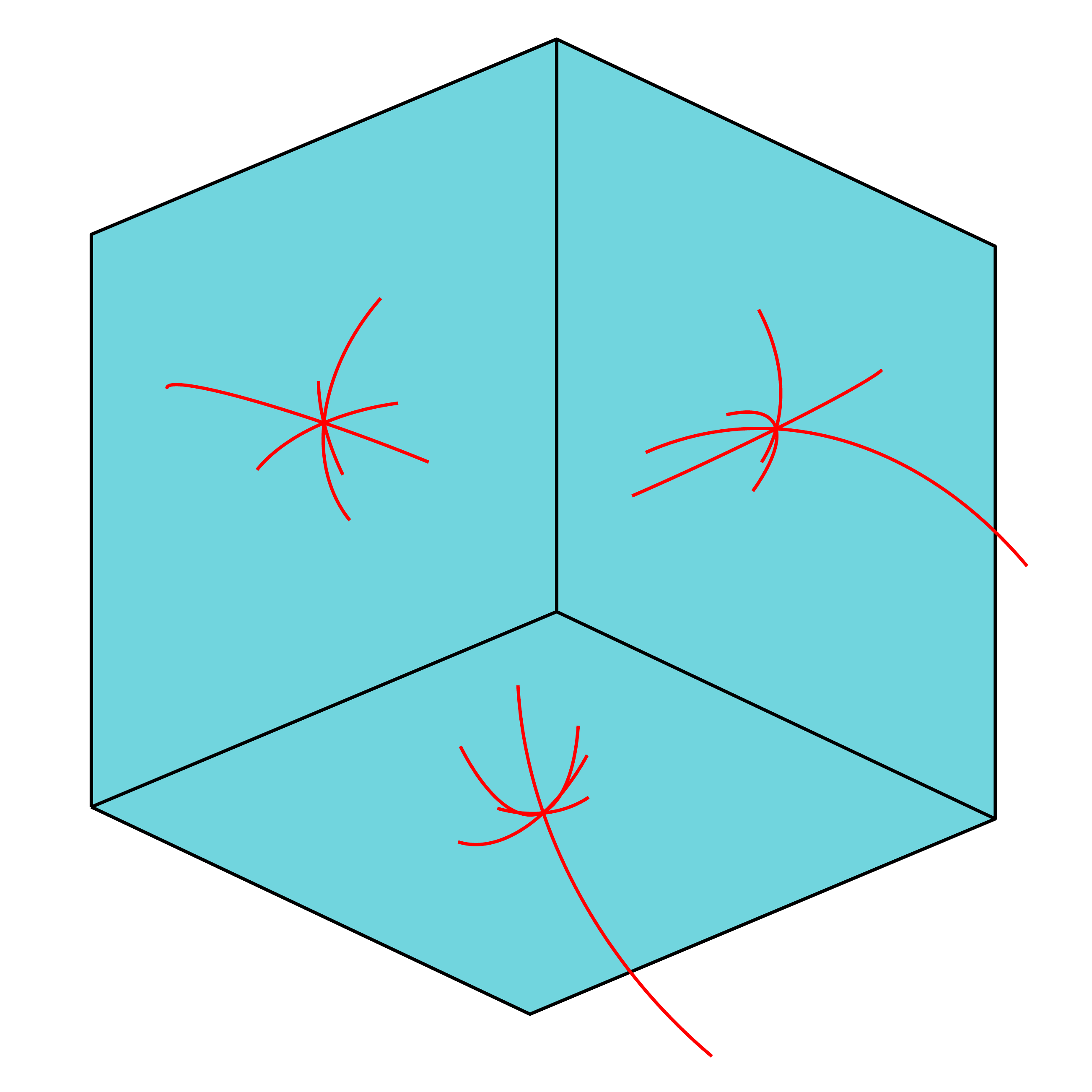}
    \caption{Circular Segments from three different walls}
    \label{fig:circular_segments_three_walls}
\end{figure}

From a triple $(\overrightarrow{v},\overrightarrow{w},\Delta)$, we can construct circular segments which intersect with the wall from which $\overrightarrow{w}$ was reflected.
As illustrated in \mbox{Fig.~\ref{fig:circular_segments_three_walls}}, circular segments from the same wall intersect if the measurements are perfect, which can be used to find reflections from the same wall. 
Note that, with measurement errors, the circular segments would not perfectly intersect, but segments from the same wall would still be close to each other, meaning that we need to identify \textit{close} segments in order to find segments belonging to the same wall.
\FloatBarrier
One approach we use to determine the closeness of two circular segments is to apply \textit{Unit Sphere Inversion} to their respective endpoints, which gives us the endpoints to a finite line segment.\\
For a given vector $\vec{v}=(x,y,z)$, we compute it's inversion $v':=\frac{(x,y,z)}{x^2+y^2+z^2}$. This definition is based on the Book "Introduction to Geometry" by \cite{coxeter}, where its also shown that inverting the points of a circle that goes through the origin results in a line.
Since all measurements are relative to the receiver, we can assume that the receiver is at the origin, which means all circular segments go through the origin.
Combining this with the previous observation from Coxeter, we know that applying the inversion to the circular segments gives us finite line segments.
By doing so, we can reduce the problem of finding close circular segments to finding close line segments.\\
Let $\ell$ be the list of all finite line segments we get from our measurements, we now want to find subsets of $\ell$ with "close" lines which will then be our measurement clusters.
We iterate over each line $l\in\ell$ and check whether there is a cluster $c$ for which the distance $d(l,c)$ is below a threshold $t$.
If $d(l,c)\leq t$, we add line $l$ to cluster $c$ and then continue with the next line.
Since we don't have any clusters in the beginning, we initialize out first cluster with the first line.\\
When computing the distance $d(l,c)$ for a given line $l$ and a cluster $c$, there are two cases:
\begin{align}\label{dlc}
    d(l,c) = 
    \begin{cases}
        d(l,l_c) &\text{if} c=\{l_c\}\\
        d(l,center_c) &\text{if} \Vert c\Vert > 1
    \end{cases}
\end{align}
The first case applies if $c$ contains only one line $l_c$. 
The second case applies if $c$ contains at least two lines.
Once we add a second line to a cluster with only one line, we also store the clusters $center_c$, which is initialized with the closest points between two lines and is adjusted if we add more lines.\\
Once all lines are added to a cluster, for each cluster of lines we get one cluster of measurements containing the respective measurements the lines were originally computed from.\\
In the worst case, we get one cluster per line meaning that our \textit{Inversion} algorithm has $\mathcal{O}(n^2)$ time complexity,  where $n$ is the number of measurements.

\subsubsection{Gnomonic Projection Approach}\label{gno}

As an alternative approach to the \textit{Unit Sphere Inversion} approach, we can also use \textit{Gnomonic Projection} to cluster the circular segments, discussed in \mbox{section \ref{inversion}}. \cite{geissler_projection} came up with this technique as part of his project. 
In the first step, we project each circular segment onto the unit sphere with radius 1, again assuming the receiver to be at the origin.
For a vector $v$, we simply compute $v':=\frac{v}{\Vert v\Vert}$ to map $v$ onto the unit sphere.
Applying this normalization to the two end points of a circular segment, we can the compute the latitude and longitude of the end points using
\begin{align}\label{arcsin}
    lat = \arcsin(z)
\end{align}
\begin{align}\label{arctan} 
    lon = \arctan(y/x)
\end{align}
Next, we choose twelve evenly spaced \emph{hemisphere center points} $h1,\dots,h_{12}$ on the unit sphere with a random rotation.
For each of these hemisphere center points, we also compute their latitude and longitude.\\
For a given hemisphere center point $h_i$, let $\varphi_{h_i}$ and $\psi_{h_i}$ be the latitude and longitude respectively.
For a given end point $p$ of an arc on the sphere, let $\varphi_p$ and $\psi_p$ be the latitude and longitude of the point.
With that, we can now compute the gnomonic coordinates of $p$ on the hemisphere around $h_i$ using
\begin{eqnarray} \label{eqxy}
    x &=& \frac{\cos(\varphi_p)\sin(\psi_p-\psi_{h_i})}{\cos(c)}\\
    y &= &\frac{1}{\cos(c)}\cdot(\cos(\varphi_{h_i})\cos(\varphi_p)-\\
    \nonumber&&\sin(\varphi_{h_i})\cos(\varphi_p)\cos(\psi_p-\psi_{h_1}))\\ 
    \nonumber\text{using}\\
    cos(c) &=& \sin(\varphi_{h_i})\sin(\varphi_p)+\\
    \nonumber&&\cos(\varphi_{h_i})\cos(\varphi_p)\cos(\psi_p-\psi_{h_1})
\end{eqnarray}

Computing the gnomonic coordinates of the two endpoints of an arc gives us the two end points on a two-dimensional plane, which we refer to as the \textit{Gnomonic Projection}.\\
We now create an \textit{intersection graph} $G$, where each line in the \textit{Gnomonic Projection} is represented by one node.
Two nodes in the intersection graph are connected iff. the respective lines are intersecting.
We then return the connected components of $G$ as our clusters.
\subsection{Computation of Wall Direction}\label{wd}

The clustering step, discussed in \mbox{section \ref{clustering}}, computes sets of measurements $(u,v,\Delta)$ such that all reflected signals in one set were reflected from the same wall, apart from some false positives.
We call these sets \textit{clusters}.\\
For each cluster, we now compute the direction and the distance of the respective wall, which is necessary to then compute the sender positions in  Section \ref{localization}.
A wall's distance and direction is encoded by its \textit{wall normal vector} which, geometrically can be defined as the vector from the receivers position to the closest point on the wall.

\subsection{Compute Wall Direction from two Measurements}\label{twomeasure}

For two given measurements $(\vec{v_1}, \vec{w_1}, \Delta_1), (\vec{v_2}, \vec{w_2}, \Delta_2)$, for which the reflections are from the same wall, we can compute the direction of the wall using the nested cross product.
\begin{align}\label{vecuspecific}
\vec{u} = \dfrac{(\vec{v_{1}} \times \vec{w_{1}}) \times (\vec{v_{2}} \times \vec{w_{2}})}{\mid (\vec{v_{1}} \times \vec{w_{1}}) \times (\vec{v_{2}} \times \vec{w_{2}})  \mid}
\end{align}
Due to the non-commutativety of the cross product, we need to either select $\vec{u}$ or $-\vec{u}$ as the wall's direction.
We do so by comparing both option both $\vec{u}$ and $-\vec{u}$ to the reflected signals $\vec{w_1},\vec{w_2}$ and select the option which minimizes the function $d : \vec{v} \mapsto \vert \vec{v} \cdot \vec{w_1} - 1 \vert + \vert \vec{v} \cdot \vec{w_2} - 1 \vert$.
Note that all vectors $\vec{u},\vec{w_1},\vec{w_2}$ have length 1.\\
To use the formula above for more than two measurements, we simply apply it to pairs of measurements and then take the average over all results.
This leads to the question which pairs to use.
For measurements $m_1,\dots,m_n$, we compare the following selections of pairs of measurements:
\begin{itemize}
    \item \textit{All Pairs}: All possible combinations of pairs of measurements. i.e. $(m_1,m_2),(m_1,m_3),\dots,(m_2,m_3),\dots,(m_{n-1},m_n)$
    \item \textit{Disjoint Pairs}: $(m_1,m_2),(m_3,m_4),\dots,(m_{n-1},m_n)$
    \item \textit{Overlapping Pairs}: $(m_1,m_2),(m_2,m_3),\dots,(m_{n-1},m_n)$
\end{itemize}
Note that using the \textit{All Pairs} method leads to a quadratic algorithm, while using the \textit{Disjoint Pairs} or \textit{Overlapping Pairs} methods gives us an algorithm with linear time complexity. These comparisons between the combinations of pairs of measurements had been carried out by \cite{grugel_wall_normal} as part of his project.

\subsubsection{Computing the distance of a Wall}\label{wall_distance}

Once we know the direction $\vec{u}$ of a given wall, we now also compute its distance.
For a given measurement $(\vec{v},\vec{w},\Delta)$, we can compute the position $\vec{s}$ of the respective sender using the \textit{Wall Direction} formula:
\begin{align} \label{spv}
    \vec{s} = p\vec{v}, 
    \end{align}
    \text{ with}\\
    \begin{align} \label{eqp}
    p = \frac{\Delta\vec{w}\cdot\vec{b}}{(\vec{v}-\vec{w})\cdot\vec{b}} 
    \end{align}
    \text{, using}\\
    \begin{align}\label{eqbdef}
    \vec{b} = (\vec{u}\times\vec{v})\times\vec{u}
\end{align}
We can then compute the distance $d:=\Vert\vec{n}\Vert$ using
\begin{align} \label{eqnvector}
    \vec{n} = \frac{\vec{s}+(p+\Delta)\vec{w}}{2}\cdot\vec{u}
\end{align}
We compute this for each available measurement and then take the average over all results as the final distance of the wall.

\section{Localization}\label{localization}

\subsection{Compute Sender Position using Single Wall}\label{singlewall}

Using a measurement cluster and respective wall normal vector as input, we have four methods for computing the position of a sender for each measurement.

\subsubsection{Map to Normal Vector}\label{mtnormal}
The \textit{Map to Normal Vector} method computes the position $\vec{s}$ of a sender using Equation (\ref{spv})
    \text{, where}\\

\begin{align} \label{eqpvectordelta}
    p = \frac{(2\vec{n}-\Delta\vec{w})\cdot\vec{n}}{(\vec{v}+\vec{w})\cdot\vec{n}}
\end{align}

\begin{figure}[htbp]
\centering
\includegraphics[width=0.8\columnwidth]{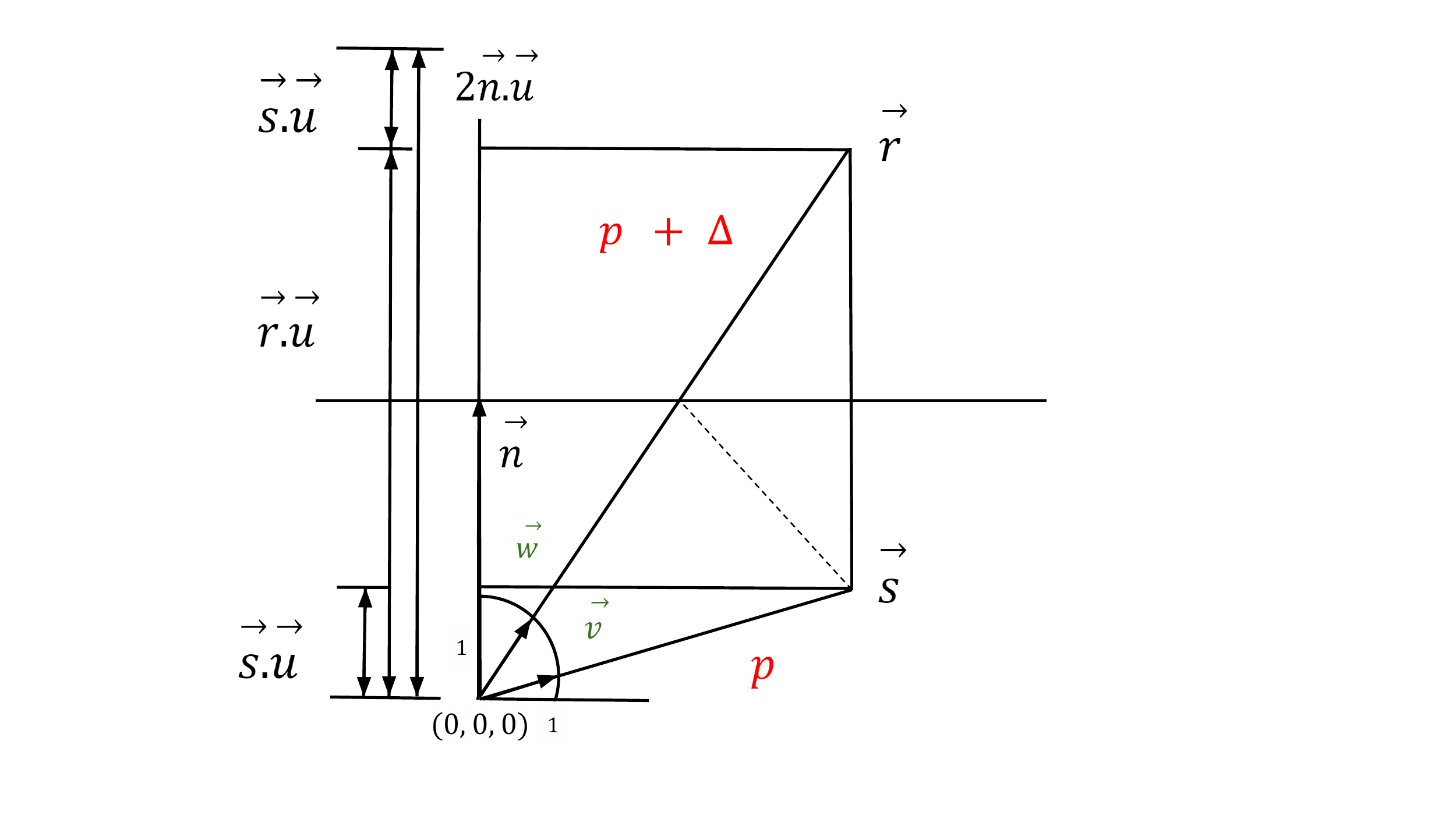} 
\caption{Visualisation of the \textit{Map to Normal} algorithm}
\label{case_map_to_normal}
\end{figure}
\FloatBarrier
\subsubsection{Reflection Geometry}\label{reflectgeometry}
The \textit{Reflection Geometry} method computes the position $\vec{s}$ of a sender using Equation (\ref{spv})
    \text{, where}\\
\begin{align}
    p = \frac{2(\vec{n}\cdot\vec{n})(\vec{w}\cdot\vec{b})}{(\vec{v}\cdot\vec{n})(\vec{w}\cdot\vec{b})+(\vec{v}\cdot\vec{b})(\vec{w}\cdot\vec{n})}
\end{align}    
    \text{, with} $\vec{b}$ defined by Equation (\ref{eqbdef}).

\begin{figure}[htbp]
\centering
\includegraphics[width=0.8\columnwidth]{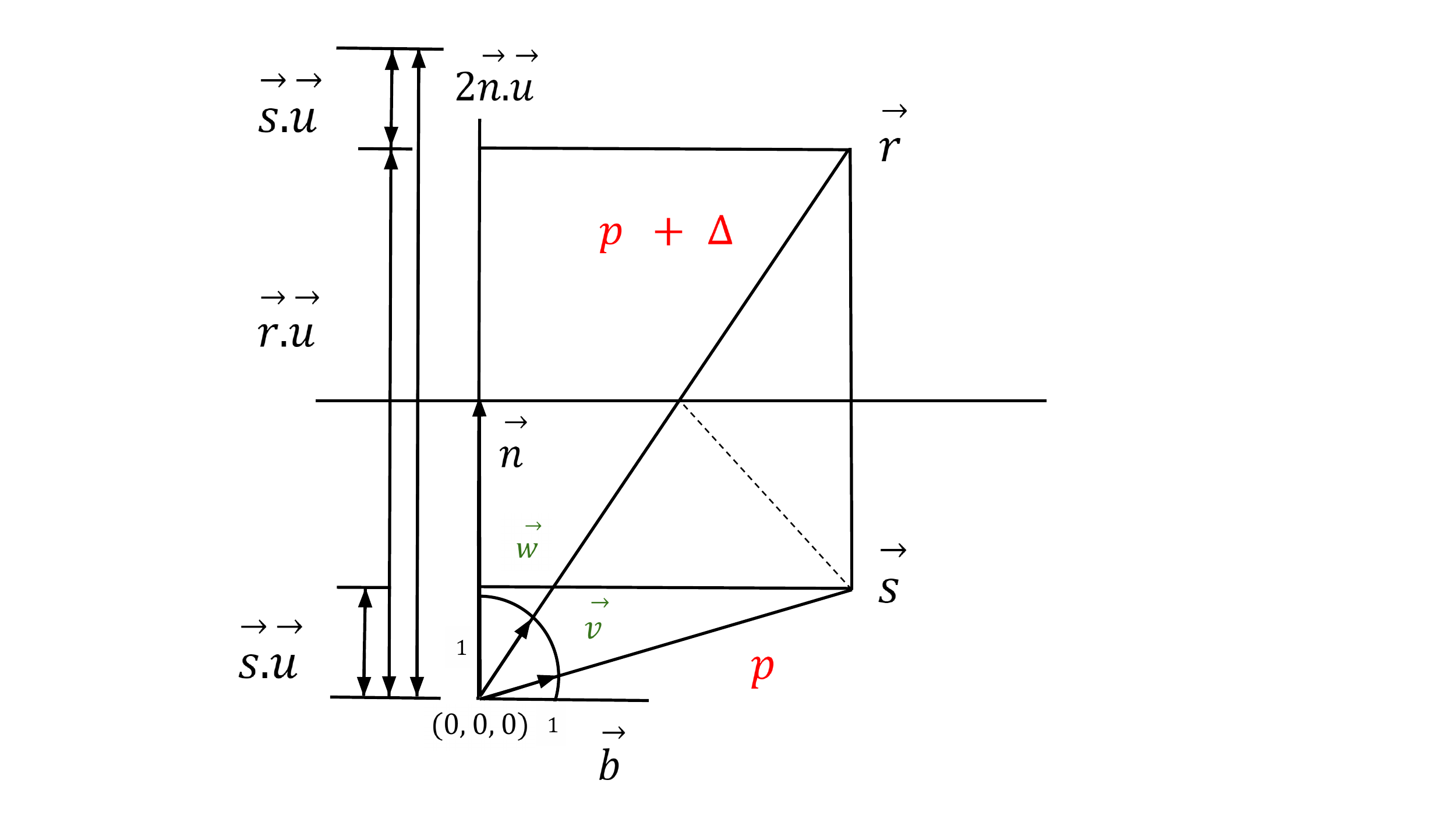} 
\caption{Visualization of the \textit{Reflection Geometry} algorithm}
\label{case_refl_geom}
\end{figure}
\FloatBarrier
\subsubsection{Wall Direction}\label{walldirectioncalc}
The formula for this method has already been introduced in \mbox{section \ref{wall_distance}}, where it is used for computing the distance of a given sender, using just the direction $\vec{u}$ of a wall in Equations \mbox{\eqref{eqp} and \eqref{eqnvector}}.

\FloatBarrier

\subsubsection{Closest Lines}\label{closestlinesapproach}

\begin{figure}[htbp]
\centering
\includegraphics[width=0.9\columnwidth]{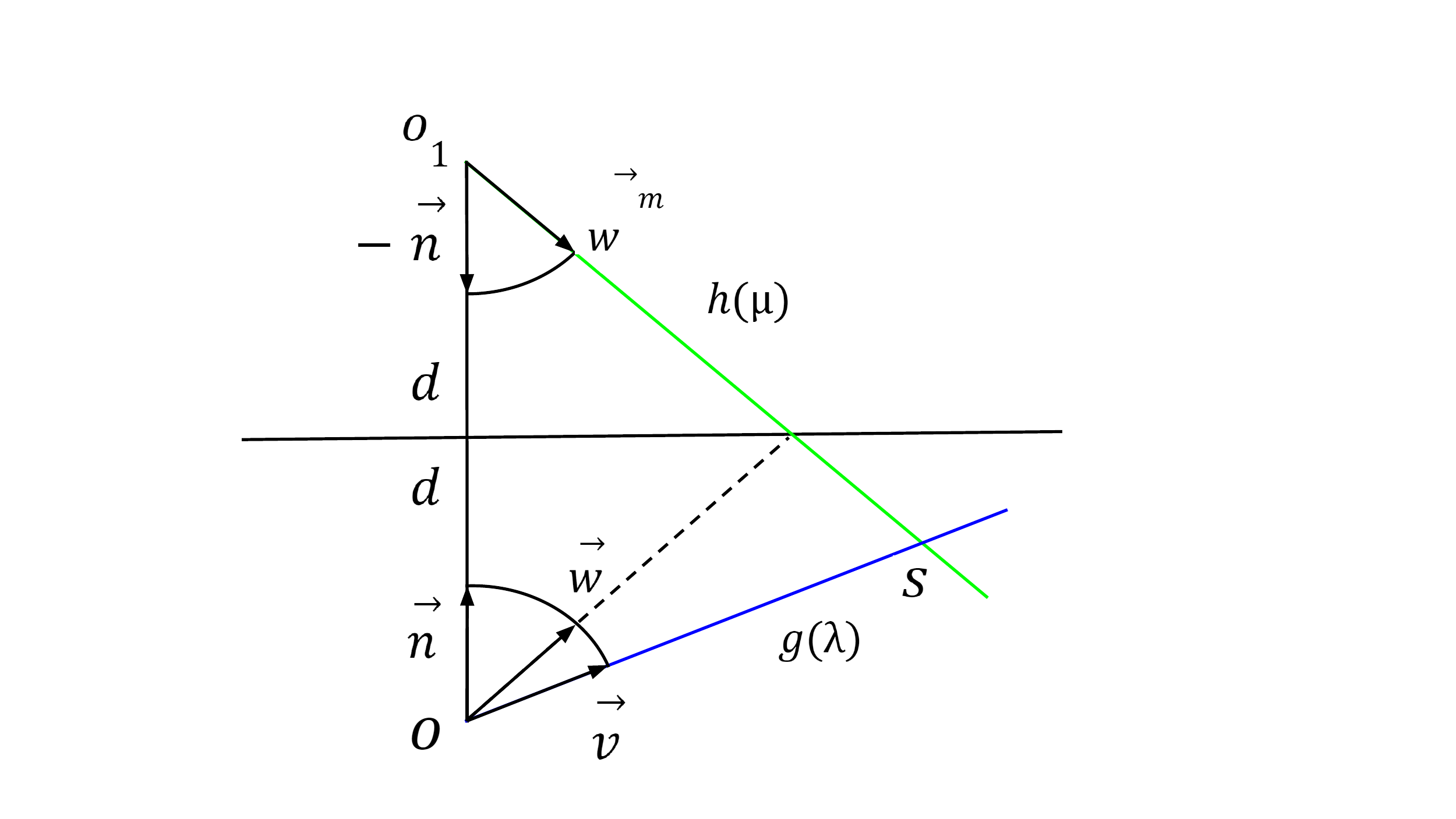}  
\caption{Visualization of the \textit{Closest Lines} algorithm}
\label{case_reflections_closest_lines}
\end{figure}

This method was introduced by \cite{weiss_localization} as part of his Bachelor Thesis. The \textit{Closest Lines} method computes a senders position by computing the closest point between two lines.
For the lines we choose
\begin{align}\label{glambda}
    g(\lambda) : x = \lambda\vec{v}
\end{align}
\begin{align}\label{hmu}
    h(\mu) : x = o_{1} + \mu \vec{w^m}
\end{align}
Where $o_{1}:= 2\vec{n}$ is the mirrored receiver position and $\vec{w^m}:=\vec{w}-2(\vec{w}\cdot\vec{n})\cdot\frac{\vec{n}}{\Vert\vec{n}\Vert}$ is $\vec{w}$ mirrored on the wall.
In \mbox{Fig.~\ref{case_reflections_closest_lines}}, line $g$ is drawn in green and $h$ is drawn in blue.

\FloatBarrier

\subsection{Wall Selection}\label{wallselection}

Note that the introduced methods to compute sender positions only use one wall, while in the clustering step we compute multiple clusters of measurements from which we can compute multiple wall normal vectors.
This leads to the question which wall to use in the localization step.
We introduce three simple methods for choosing a wall.
\textit{Largest Cluster} will simply use the wall where the respective measurement cluster contains the most elements.
For a different method titled \textit{Narrowest Cluster}, we compute the average angular distance of the reflected signals to the wall normal vector that is generated using the respective cluster.
We then choose the cluster which minimizes this value.
For our third method, \textit{Unweighted Average}, we simply compute each sender position once for each wall and then take the average position for each sender as the final output.

\subsection{Using Multiple Walls for Localization}\label{multiwalls}

Another approach to sender localization we introduce is to use multiple walls in one algorithm.
This algorithm titled \textit{Closest Lines Extended} is based on the \textit{Closest Lines} approach.
For a given direct signal, we consider all the respective reflected signals which were assigned in the clustering step, which means for each of the reflected signals we have one wall normal vector.
We then take a line $g$ like with \textit{Closest Lines}  and then we also use one additional line for each pair of wall normal vector and reflected signal, using the same formulas as in the regular \textit{Closest Lines} approach.
We then use an algorithm introduced by \cite{han-bancroft-closest-lines} to compute the closest point between all the lines and then use that as our final sender position.

\begin{figure}[htbp]
\centering
\includegraphics[width=1.1\columnwidth]{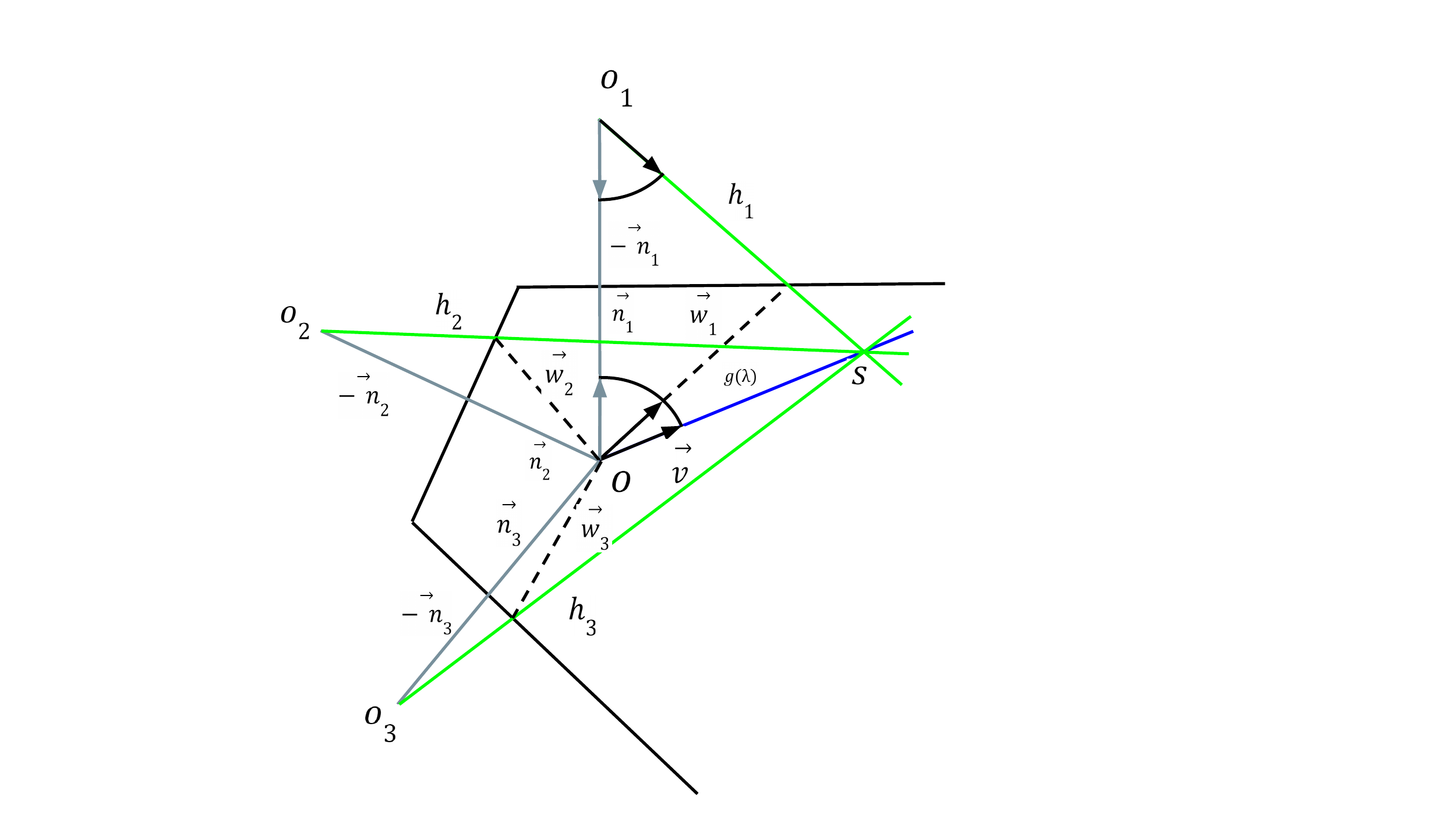}  
\caption{Visualization of the \textit{Closest Lines Extended} algorithm}
\label{case_reflections_closest_lines-extended}
\end{figure}
\FloatBarrier

\section{Simulation Setup}\label{simulation_setup}
For testing the presented algorithms, we first generate simulated measurements using 20 randomly places sender positions in a $2\times2\times2$ meters cube. We set the receiver position to be exactly at the center of the room.
For simulating the data, we assume that the receiver receives one direct signal from each sender and one direct sender from each wall for each sender.
We assume the of angle of incidence to be equal to the exit angle for the reflected signals.\\
To get a more realistic simulation of input signals, we apply three types of errors the simulated data.\\
Firstly, we use a von Mises distribution to alter the angle of all direct and reflected signals $\vec{v}$ and $\vec{w}$. For the von Mises distribution, we use a concentration value $\kappa=131.312$.
We also use a normal distribution with a standard deviation of $10cm$ to simulate errors on the $\Delta$ values. If this results in a negative $\Delta$, we take the absolute value instead.
In addition to the simulated error on the measurements itself, we also randomly assign $5\%$ of reflected signals to a different direct signal.

\section{Simulation Results}\label{simulation_res}
Using the simulated input data, as described in \mbox{section \ref{simulation_setup}}, we now run all possible combinations of algorithms on the same inputs. 
For each computed sender position, we save the euclidean distance to the actual position of the sender that the respective direct signal was emitted from.
In the following, we will refer to this euclidean distance as the \textit{offset}.
We ran 500 experiments, each with new random sound source positions, and then analyse all the offset values from all experiments.\\

Note that, when the \textit{Closest Lines Extended} algorithm is used, there is no need to choose one of the computed walls, since \textit{Closest Lines Extended} uses all walls.
Therefore, all combinations of algorithms using the \textit{Closest Lines Extended} algorithm will implicitly use the \textit{Unweighted Average} method for wall selection.\\

\begin{figure}[htbp]
\centering 
\includegraphics[width=1.0\textwidth]{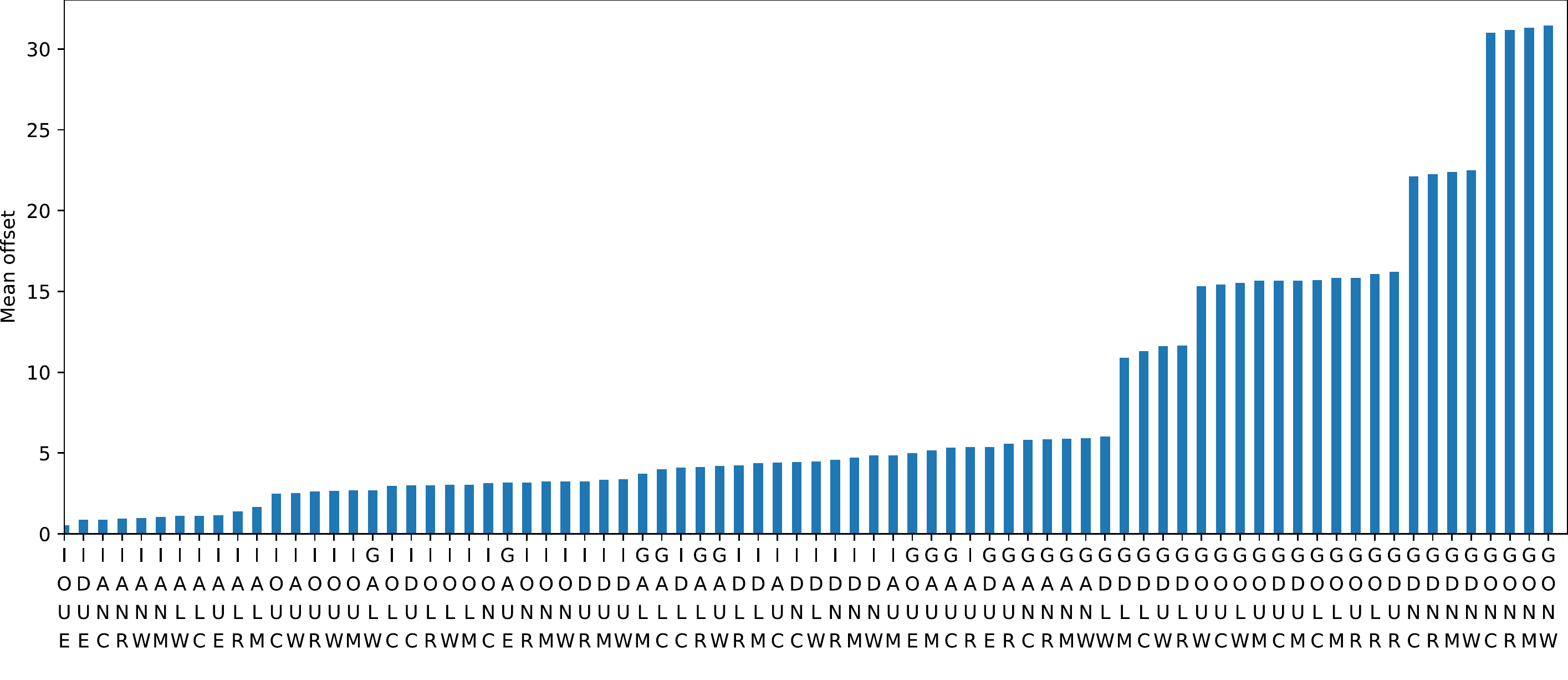}  
\caption{All 78 combinations of algorithms sorted according to increasing mean offset. }
\label{meansorting}
\end{figure}

\begin{figure}[htbp]
\centering
\includegraphics[width=1.0\columnwidth]{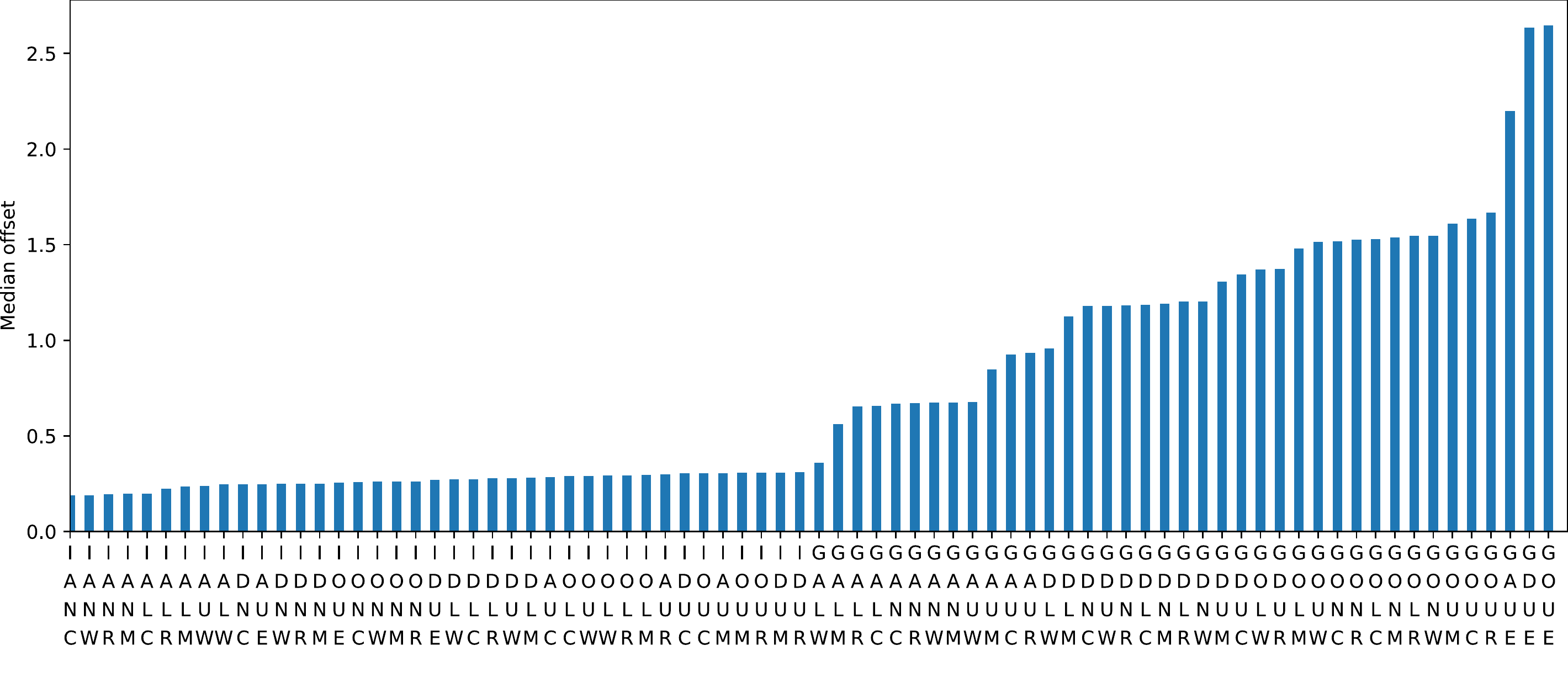} 
\caption{All 78 combinations of algorithms sorted according to increasing median offset. }
\label{mediansorting}
\end{figure}

\begin{figure}[htbp]
\centering
\includegraphics[width=1.0\columnwidth]{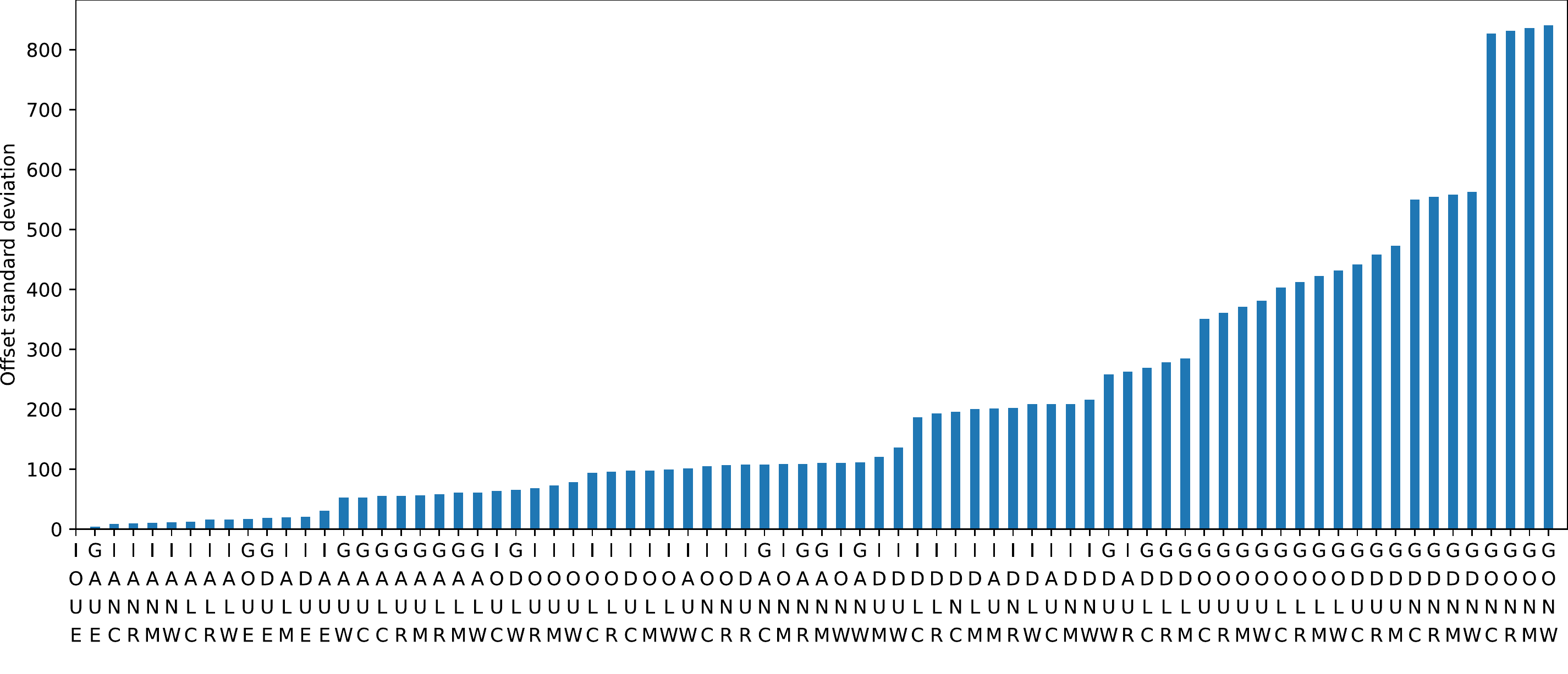} 
\caption{All 78 combinations of algorithms sorted according to increasing values of standard deviation of offset}
\label{stdsorting}
\end{figure}

For readability reasons, we use abbreviations for the algorithm names in \mbox{Fig.~\ref{meansorting} to \ref{stdsorting}}. I stands for \textit{Inversion}, G for \textit{Gnomonic Projection}, A for \textit{All Pairs}, O for \textit{Overlapping Pairs}, D for \textit{Disjoint Pairs}, U for \textit{Unweighted Average}, N for \textit{Narrowest Cluster}, L for \textit{Largest Reflection Cluster}, C for \textit{Closest Lines}, W for \textit{Wall Direction}, R for \textit{Reflection Geometry}, E for \textit{Closest Lines Extended} and M for \textit{Map to Normal Vector}.\\

The \mbox{Fig.~\ref{meansorting}} represents all combinations of algorithms sorted in increasing order of mean offset values. This is also verified using the box-whisker plot, \mbox{Fig.~\ref{best_by_mean}} which shows a similar output but for the first five combinations of most accurate algorithms with regards to mean offset.

\FloatBarrier
The \mbox{Fig.~\ref{mediansorting}} represents all combinations of algorithms sorted in increasing order of median offset values. This is also verified using the box-whisker plot, \mbox{Fig.~\ref{best_by_median}} which shows a similar trend but is more specific to the first five most accurate combinations of algorithms sorted by median offset.

\FloatBarrier
The \mbox{Fig.~\ref{stdsorting}} represents all combinations of algorithms sorted in increasing order of standard deviation of the offset values.

\FloatBarrier
The same results as above can be seen also here using the following tabular form where red crosses stand for mean, blue crosses are for median and green crosses signify standard deviation of offsets. These have been arranged from left to right in the table in increasing order of their sorted values.

\begin{figure}[htbp]
\centering
\includegraphics[width=0.95\textwidth]{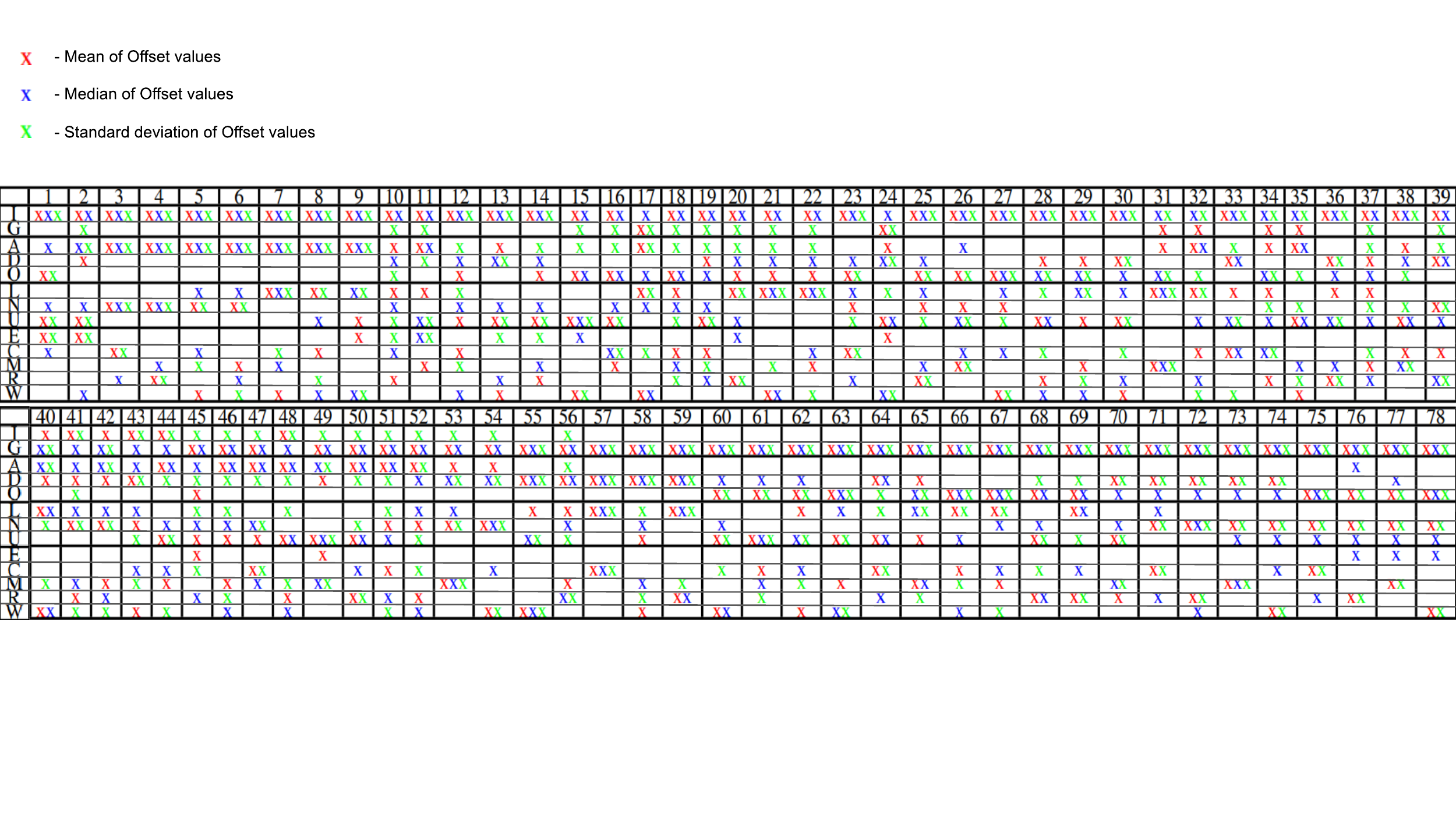} 
\caption{The cumulative tabular form of the above three graphs,i.e; 
 \mbox{Fig.~\ref{meansorting}},\mbox{Fig.~\ref{mediansorting}} and \mbox{Fig.~\ref{stdsorting}}. The rows signify the algorithm type such as I for \textit{Inversion}, D for \textit{Disjoint Pairs} etc. and the columns indicate all the 78 combinations of these algorithms in increasing order by mean, by median and by standard deviation of offsets, depending on their respective color schemes, shown together.}
\label{tabularrep}
\end{figure}
\FloatBarrier

\subsection{Most Accurate Combination of Algorithms based on median Offset}\label{median_combo_offset}

\begin{figure}[htbp]
\centering
\includegraphics[width=0.8\columnwidth]{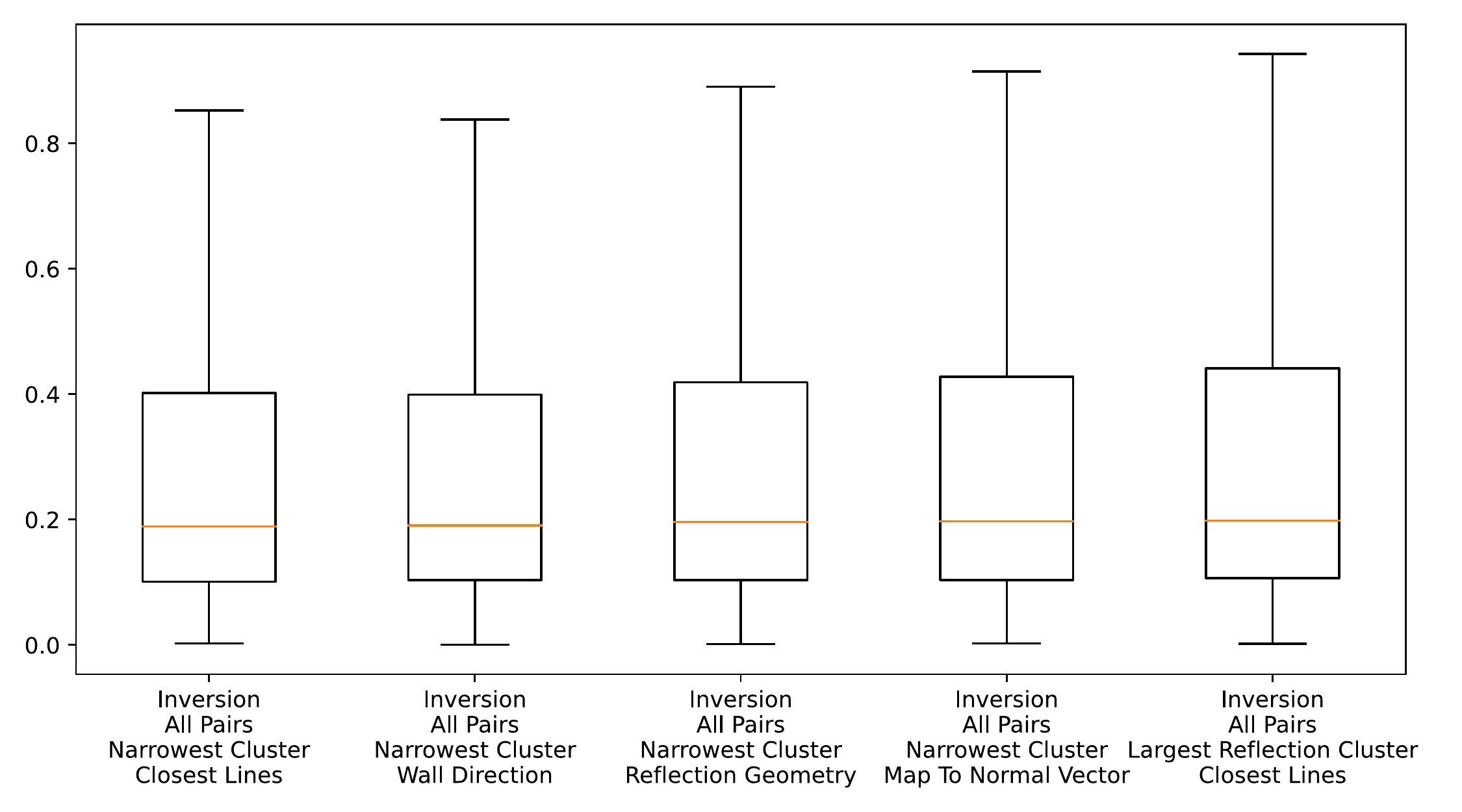} 
\caption{The 5 combinations of algorithms with lowest median offset. 
The orange Line shows the median offset. 
The upper and lower bounds of the boxes indicate the first quartile $Q_1$ and the third quartile $Q_3$ of the data points.
The lower and upper whiskers are placed at $Q_1-1.5\cdot IQR$ and $Q_3 + 1.5\cdot IQR$ respectively, where $IQR=Q_3-Q_1$.
Note that outliers outside of the whiskers are not shown in this graph.
The first line of the x-axis labels shows the clustering algorithm used, the second line specifies the averaging method used for wall direction computation, the third line shows the wall selection method and the fourth line shows the sender localization method. 
}
\label{best_by_median}
\end{figure}

\mbox{Fig.~\ref{best_by_median}} presents the offsets of the five algorithms with the lowest median offset using a box-whisker plot. 
The plot shows that the combination of the \textit{Inversion } clustering method and the \textit{All Pairs} method for wall direction computation yields the most accurate results, judging by the median offset. For wall selection, the \textit{Narrowest Cluster} method, combined with the \textit{Closest Lines} algorithm for localization, gives the most accurate results.

\FloatBarrier
\subsection{Most Accurate Combination of Algorithms based on mean Offset}\label{mean_combo_offset}
\begin{figure}[htbp]
\centering
\includegraphics[width=0.8\columnwidth]{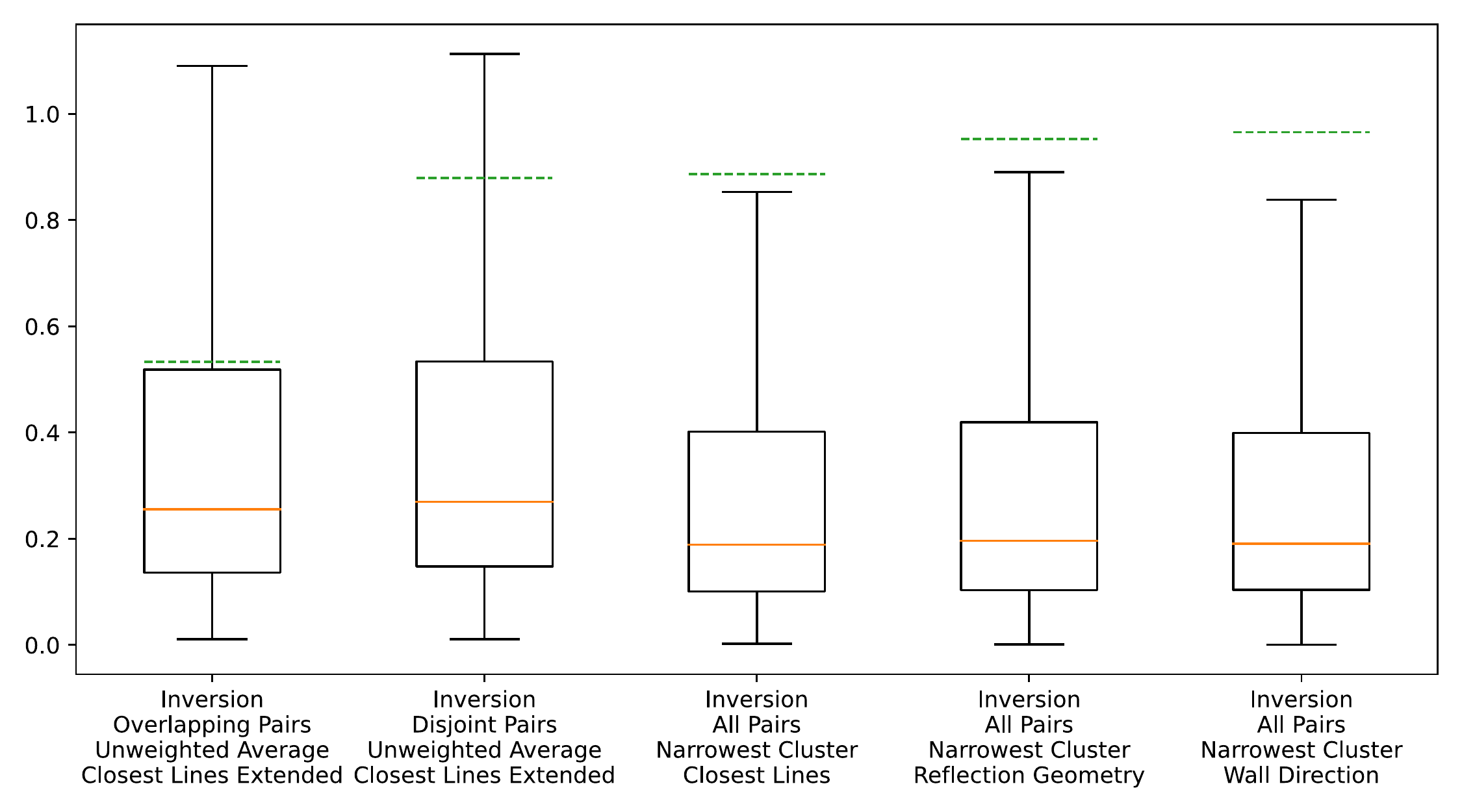} 
\caption{The 5 combinations of algorithms with lowest mean offset. 
The boxes and whiskers are the same as in \mbox{Fig.~\ref{best_by_median}}, the added green dashed line indicates the respective mean values.
Here, the five combinations with the least mean are shown.
}
\label{best_by_mean}
\end{figure}

\mbox{Fig.~\ref{best_by_mean}} shows the 5 most accurate combinations of algorithms judging by the average offset. 
Note that the mean offset, indicated by the green line, for all combinations is significantly higher than the orange line indicating the respective median.
This can be attributed to outliers with large offset values.\\
Compared to \mbox{Fig.~\ref{best_by_median}}, the \textit{Closest Lines Extended} algorithm produces lower average offset values compared to \textit{Closest Lines} algorithm using a single wall.

\FloatBarrier

\section{Conclusions and Future Work}\label{conclusionfw}

In this paper, we illustrate various techniques to compute the distance, $p$ between sender and receiver depending on various configurations of sender-receiver pairs with respect to the wall and the normal vector to it. We also cover various techniques of grouping multiple walls together using clustering algorithms of projection and inversion. In the final results we compare multiple algorithms combining several of the aforementioned techniques and come to the conclusion that the combinations involving \textit{Inversion} perform significantly better than those with \textit{Gnomonic Projection} as can be verified from the mean and median bar as well as box-whisker plots for the offsets in \mbox{section \ref{simulation_res}}.\\
As a possible future work, we would like to move forward into employing the above techniques on real-world experimental data instead of simulations as have been currently used here. We would like to do so since it is not clear how comparable our simulated data is especially in regards to the simulated errors, compared to data from real-world experiments.
\bibliographystyle{IEEEtran}
\bibliography{SimulationOfPrototypical3D}

\end{document}